\documentclass[letterpaper, 10 pt, conference]{ieeeconf}  

\IEEEoverridecommandlockouts                              

\usepackage{graphics} 
\usepackage{subfigure}
\usepackage{epsfig} 
\usepackage{amsmath} 
\usepackage{amssymb}  
\usepackage{booktabs}
\usepackage{algorithm}
\usepackage{algpseudocode}
\usepackage{cite}
\usepackage{url}
\usepackage{xcolor}
\usepackage[outdir=./]{epstopdf}
\title{\LARGE \bf
Online estimation of the hand-eye transformation from surgical scenes}

\author{Krittin Pachtrachai, Francisco Vasconcelos, and Danail Stoyanov
\thanks{Krittin Pachtrachai, Francisco Vasconcelos, and Danail Stoyanov are with the Wellcome / EPSRC Centre for Interventional and Surgical Sciences (WEISS) and the Department of Computer Science, University College London, Gower Street, WC1E 6BT, UK.
        {\small \{krittin.pachtrachai.13, f.vasconcelos, danail.stoyanov\}@ucl.ac.uk}}%
}

\begin{document}

\maketitle
\thispagestyle{empty}
\pagestyle{empty}

\bstctlcite{BSTcontrol}

\begin{abstract}
Hand-eye calibration algorithms are mature and provide accurate transformation estimations for an effective camera-robot link but rely on a sufficiently wide range of calibration data to avoid errors and degenerate configurations. To solve the hand-eye problem in robotic-assisted minimally invasive surgery and also simplify the calibration procedure by using neural network method cooporating with the new objective function. We present a neural network-based solution that estimates the transformation from a sequence of images and kinematic data which significantly simplifies the calibration procedure. The network utilises the long short-term memory architecture to extract temporal information from the data and solve the hand-eye problem. The objective function is derived from the linear combination of remote centre of motion constraint, the re-projection error and its derivative to induce a small change in the hand-eye transformation. The method is validated with the data from da Vinci Si and the result shows that the estimated hand-eye matrix is able to re-project the end-effector from the robot coordinate to the camera coordinate within 10 to 20 pixels of accuracy in both testing dataset. The calibration performance is also superior to the previous neural network-based hand-eye method. The proposed algorithm shows that the calibration procedure can be simplified by using deep learning techniques and the performance is improved by the assumption of non-static hand-eye transformations.

\end{abstract}

\section{INTRODUCTION}
\label{sec:intro}

Robotic-assisted minimally invasive surgery (RMIS) is currently utilised in multiple interventional settings, including prostatectomy and colorectal surgeries. In this procedure, a robot handling a laparoscopic camera and articulated instruments is tele-operated by a surgeon sitting at a console \cite{9805581}. In the current setting, RMIS still has little computer assisted intervention (CAI) with respect to its full potential. Further improvements could include augmented reality to enhance visualisation and localisation of sub-anatomical information \cite{augmented-reality}, and virtual fixtures to provide force feedback according to the instruments' movement and the surrounding \cite{6634270}. One of the main current challenges towards these CAI applications involves the spatial representation of all relevant elements (laparoscope, instruments, anatomy) under a single reference frame. The pose of all robot articulations can be computed from the configuration of its joints (forward kinematics). On the other hand, the target anatomy in the operative site is only visualised through a laparoscope. Aligning the coordinate systems of the robot articulations and its camera is conventionally called hand-eye calibration \cite{Lenz}.

The conventional procedure for hand-eye calibration involves tracking the motion of a camera attached to a robot through a calibration target (typically a checkerboard). The method is considered to yield accurate calibration results for a wide range of robotic applications, however, this is not the case for RMIS \cite{Pachtrachai2018}. For hand-eye calibration to be accurate, its camera motion needs to explore all its 6 degrees of freedom in sufficient range \cite{Lenz} to mitigate the influence of measurement noise. In RMIS, robot motion is mechanically confined around a remote centre of motion (RCM) \cite{8744556} which makes this requirement difficult to achieve. Furthermore, the conventional calibration is an off-line process that assumes the hand-eye transformation will remain constant during robot operation. This is not necessarily true in RMIS where surgical instruments are tendon-driven and their measured position can be affected by systematic offsets due to joint hystheresis and external forces. Additionally, the calibration procedure itself can be disruptive to the surgical workflow, as it requires a certain amount of planning and time before the operation, as well as the sterilisation of additional objects in the operating room (calibration targets).

In this paper we propose a method that addresses the above challenges using a deep learning based calibration approach. Our method performs online hand-eye calibration from real robotic surgery video footage, directly using the surgical instruments as calibration targets. It is also capable of dynamically updating the calibration with new measurements. While there are previous works targeting some of these challenges, we introduce the following contributions:
\begin{figure*}[!t]
\centering
\subfigure[\label{fig:dvrk-frames}]{\includegraphics[width = 0.33\textwidth]{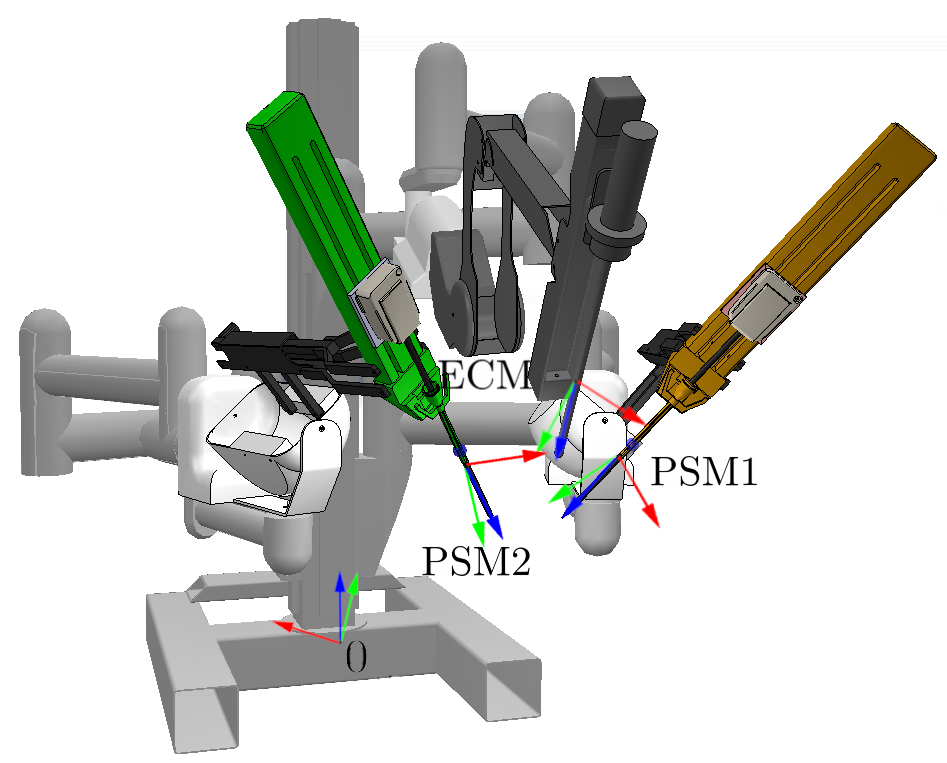}}
\subfigure[\label{fig:camera-frame}]{\includegraphics[width = 0.65\textwidth]{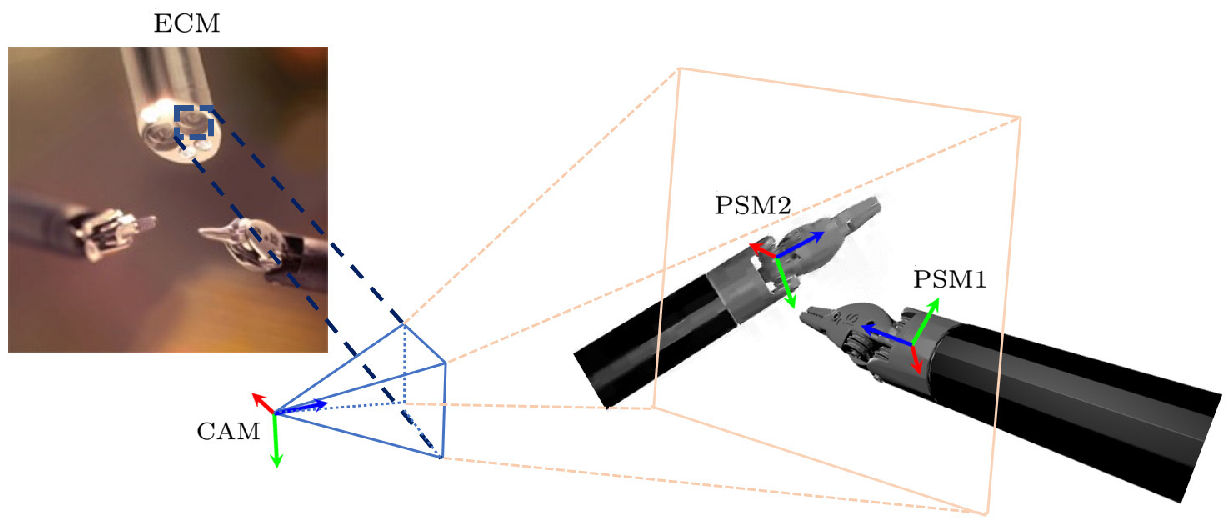}}
\caption{(a) The simulation of the da Vinci system in CoppeliaSim designed in \cite{8487187}. PSMs (Patient Side Manipulators) and ECM (Endoscope Control Manipulator) arms are connected to the robot base $0$ operating in the same kinematic chain. (b) The simulated view of the PSM arms captured from the optical centre of the laparoscope.}
\end{figure*} 

\begin{itemize}
\item We propose a novel differentiable loss function for this problem that explicitly considers dynamically varying hand-eye transformations during back-propagation. Therefore, our trained network can adapt to changes in calibration (geometric errors) and more complex kinematic errors (non-geometric errors), caused by stress and strain from tendon driven actuators due to forces at the end-effector.
\item We demonstrate that a model trained with our loss function outperforms a baseline that is trained assuming fixed hand-eye transformation. Surprisingly, our method also outperforms the baseline when the hand-eye transformation is constant, and we also provide a reasoning for this.
\item When compared to previous approaches, we further automate the calibration process by integrating it with visual tracking of instrument joints via the DeepLabCut (DLC) package \cite{DLC}.
\end{itemize}

\section{Related work}\label{sec:lite}
Conventionally, the hand-eye transformation is a rigid homogeneous mapping $\mathbf{X}$, that satisfies $\mathbf{AX} = \mathbf{XB}$, where $\mathbf{A}$ and $\mathbf{B}$ are homogeneous relative motions of the "eye" (camera) and "hand" (robot) respectively. Most approaches to hand-eye calibration involve stacking multiple instances of this equation, where $\mathbf{A}$, $\mathbf{B}$ are known measurements and we are solving for $\mathbf{X}$. Different methods formulate homogeneous transformations with different parameterisations \cite{Lenz, Pachtrachai2018, ProcrutesNPoint}. Some works also introduce additional optimisation criteria, such as epipolar constraints \cite{Abed2}, or point re-projection error \cite{AbedRadial}. However, if "hand" and "eye" motions do not satisfy certain criteria, its standard equation becomes degenerate and results in unstable calibration \cite{Huang2018OnGM}. This is a significant problem in RMIS, where RCM constraints are close enough to the degenerate case. An alternative formulation of the hand-eye equation has been proposed to handle the specific case of RCM constrained motion \cite{8744556}.

All the above mentioned methods rely on tracking the camera motion with a known calibration object, usually a checkerboard. A few works directly estimate hand-eye calibration from arbitrary static scenes, relying on Structure-from-Motion (SfM) \cite{Linfinity}. However, SfM is a computationally heavy offline technique, and does not handle dynamic non-rigid scenes present in RMIS applications. A more promising approach in this context is to visually track surgical robot instruments and use them as the hand-eye calibration targets \cite{8004522}. While generally more reliable than checkerboard calibration, these methods propagate any tracking errors to the estimated calibration. The most recent work in \cite{9781386} requires sampling points along a surgical instrument from an edge detection. When tracking instruments against a clear background, the algorithm delivers around 1 mm calibration error in average. However, this is an offline procedure that requires a preoperative setup which is still disruptive to the normal workflow in an operation room.

Deep learning techniques have been widely used for visual recognition tasks \cite{10.1145/3065386, DBLP:journals/corr/abs-1907-05272}, and this also includes sensor calibration problems \cite{calibRCNN}. In the context of hand-eye calibration in RMIS, \cite{9495280} proposes a solution based on a LSTM model, validated on the da Vinci Surgical robot. However, it only utilises a few frames where the end-effector positions are manually marked, and its loss function implicitly assumes a constant hand-eye transformation. Therefore, this approach does not take into account errors from the robot kinematic chain.  

\begin{figure*}[!t]
\centering
\label{fig:LSTM-calib}\includegraphics[width = 1.0\textwidth]{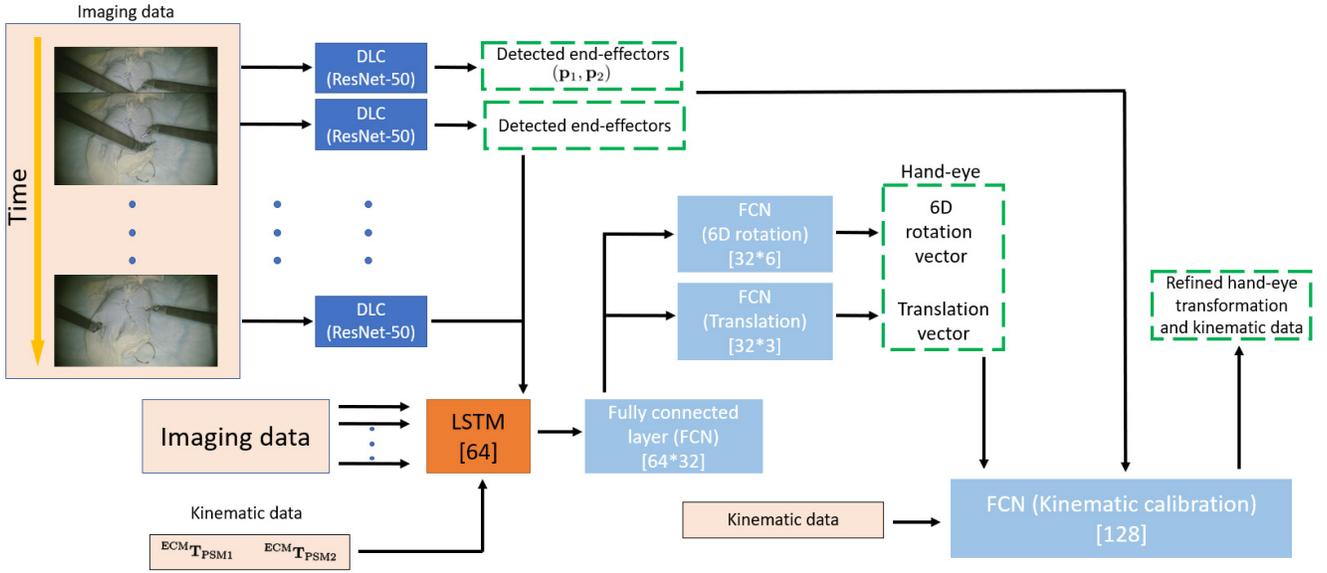}
\caption{The architecture of the proposed network to solve the hand-eye problem and update the parameters in RMIS using DLC \cite{DLC}, LSTM \cite{10.1162/089976600300015015} and robot calibration network. The inputs to the network are a sequence of images and the poses of the surgical instrument viewed from the camera, $^\text{ECM}\mathbf{T}_{\text{PSM1}}$ and $^\text{ECM}\mathbf{T}_{\text{PSM2}}$. The hand-eye network is defined as the three fully-connected layers following LSTM, and the robot calibration network which compensates the non-geometric error in the kinematic consists of three levels of fully-connected layer with LeakyRelu as the activation function. The dimensions of the network are $46\times64$, $64\times64$ and $64\times18$. }
\end{figure*} 

\section{METHODS}
\label{sec:methods}
We consider the da Vinci surgical system, with reference frames attached to the camera arm end-effector (ECM) and the instrument arm end-effectors (PSM1, PSM2) as shown in Figure \ref{fig:dvrk-frames}. The transformations between these reference frames can all determined from the kinematic chain of the robot joints, i.e. The kinematic data ($^\text{ECM}\mathbf{T}_{\text{PSM1}}$, $^\text{ECM}\mathbf{T}_{\text{PSM2}}$) can be computed from the robot joints. An additional camera frame (CAM) is defined at the optical centre of the laparoscope (Figure \ref{fig:camera-frame}), however, its transformation to any of ECM, PSM1, PSM2 cannot be determined from robot kinematics alone. We define our hand-eye calibration problem as the estimation of the transformation $^\text{CAM}\mathbf{T}_{\text{ECM}}$ connecting ECM and CAM using imaging and kinematic data.

We solve this using the deep learning network shown in Figure \ref{fig:LSTM-calib}. The instrument end-effector positions ($\textbf{p}_1$,$\textbf{p}_2$) are tracked from imaging data using the DLC network \cite{DLC}, which is based on ResNet-50. We manually label the end-effector positions using the toolbox provided in the paper and pre-train the network for 150,000 iterations. The output of the network are the detected positions and the uncertainty score of each detection. Note that image resizing affects the quality of the detection as the features in images are discarded. In this paper, we empirically determine that all images should be resized to $640\times360$ from $1920\times1080$ to make the training more viable and can still retain the accuracy of the detection.

To combine imaging and kinematic data, the images are converted to grayscale format and reshape into a vector before stacking with the transformations. The data is then normalised. Sequences of both imaging and kinematic data are fed into an LSTM layer to process temporal features. The output of LSTM is provided to a sequence of fully connected layers with LeakyRelu activations that output the rotation (in its 6D vector parametrisation \cite{DBLP:journals/corr/abs-1812-07035}) and translation of the hand-eye transformation $^\text{ECM}\mathbf{T}_{\text{CAM}}$. Finally, $^\text{CAM}\mathbf{T}_{\text{ECM}}$ is further refined by passing it through an 8-layer FCN, together with kinematic data and detected end-effector positions. 

\subsection{Loss function}\label{subsec:loss}

\textbf{Z-axis loss:} The work in \cite{9495280} suggests that one of the axes (typically Z-axis) assigned to the camera frame is always parallel with one of the axes at the ECM frame, due to the constrained configuration in RMIS. With this assumption, the relative motion estimation has been improved. Let $\vec{r}_z$ and $\vec{z}$ be the Z-axis of the hand-eye transformation and the Z-axis of ECM orientation, respectively. The loss can be written as follows.
\begin{equation}
\mathfrak{L}_{\text{z}} = ||\text{arccos}(\vec{r}_z^T\vec{z})||
\label{eq:rcm}
\end{equation}

\textbf{Re-projection:} Once an estimate of $^\text{ECM}\mathbf{T}_{\text{CAM}}$ is available, the positions of the end-effector of PSM1 and PSM2 can also be mapped to the camera frame. Therefore, if this estimation is accurate, the re-projected positions in the scene and the detected positions from DLC should be small. This can be formulated as the re-projection error. 
\begin{equation}
\mathfrak{L}_{\text{proj}} = \sum_{j=1}^2\sum_{i = 1}^N w_j\lvert\lvert\mathbf{p}_{1, j} - f(^{\text{CAM}}\mathbf{T}_{\text{PSM}j})\rvert\rvert
\label{eq:reproj}
\end{equation}
where we define the transformations $^{\text{CAM}}\mathbf{T}_{\text{PSM}j}={^\text{CAM}\mathbf{T}_{\text{ECM}}} {^{\text{ECM}}\mathbf{T}_{\text{PSM}j}}$. $\mathbf{p}_{1, i}$ and $\mathbf{p}_{2, i}$ are the points detected by DLC and $f(\cdot)$ is the re-projection of apoint into a camera pinhole model including its pose, intrinsic parameters and distortion coefficients (known from calibration). The term $w_j$ is the certainty score ranging from $0$ to $1$ describing the accuracy of each point detection which is an output of DLC. 

\textbf{Differentiation of re-projection:} Since the arms are typically moved around in the operating space in RMIS, the transformations are usually updated accordingly through forward kinematics. However, due to complex and highly non-linear errors in the kinematic chain caused by joint compliance or stress and strain from tendon-driven joints, the hand-eye transformation should be able to update itself and compensate the discrepancy. This cost function is a derivative of the re-projection error with respect to time.
\begin{eqnarray}
\begin{split}
\mathfrak{L}_{\text{diff}} = \sum_{j=1}^2\sum_{i = 1}^N w_j\lvert\lvert\dot{\mathbf{p}}_{1, i} - \dot{f}(^{\text{CAM}}\mathbf{T}_{\text{PSM}j})^{\text{CAM}}\dot{\mathbf{T}}_{\text{PSM}j}\rvert\rvert \\
\text{where}\quad^{\text{CAM}}\dot{\mathbf{T}}_{\text{PSM}j} = ^{\text{CAM}}\dot{\mathbf{T}}_{\text{ECM}}\ ^{\text{ECM}}\mathbf{T}_{\text{PSM}j} + \\ \ ^{\text{CAM}}\mathbf{T}_{\text{ECM}}\ ^{\text{ECM}}\dot{\mathbf{T}}_{\text{PSM}j}
\label{eq:reproj-diff}
\end{split}
\end{eqnarray}
where $\dot{\mathbf{p}}_{1, i}$ and $\dot{\mathbf{p}}_{2, i}$ are the changes of detected positions and $\dot{f}(\cdot)$ is the derivative of the re-projected function with respect to time representing the change between frames. The chain rule can be used further to derive the following equation which takes into account the changes in both the hand-eye parameters $^{\text{CAM}}\dot{\mathbf{T}}_{\text{ECM}}$ and the instruments' positions $^{\text{ECM}}\dot{\mathbf{T}}_{\text{PSM}j}$. To approximate the derivative terms in the equation, we use finite difference method to interpolate the change between the two consecutive data, assuming that the small change between them is linear.

The loss function of the network is a linear combination of the three functions $\mathfrak{L} = c_1\mathfrak{L}_{\text{z}} + c_2\mathfrak{L}_{\text{proj}} + c_3\mathfrak{L}_{\text{diff}}$ where $c_1, c_2$ and $c_3$ are empirically determined to be $100, 1.0$ and $0.75$ respectively.

For the training, we train the network for $50$ epochs using Adam optimiser with the learning rate of $1e-4$ for the hand-eye network and $1e-3$ for the robot calibration network. The learning rate is decreased by $0.75$ for every $10$ epochs. There are two datasets we use in the training and validating, the glove suturing dataset (10,000 images for training and 3,000 images for validation) and the prostatectomy dataset (9,000 images for training and 2,000 images for validation). The datasets are trained together.

\section{Experiments and results}\label{sec:exp}
The proposed algorithm is compared with the previously developed hand-eye method in \cite{9495280} in terms of the accuracy of re-projected end-effectors and the convergence rate. Two sets of data are used in the validation, where one dataset has a stationary configuration of ECM arm (glove suturing) and the other has a moving camera during prostatectomy. The latter contains more occlusion and unclear features and also the scenes where the end-effectors are not fully visible, i.e. partially or entirely out of the scene.

\begin{figure*}[!t]
\centering
\subfigure[\label{fig:reproj-glove-1}]{\includegraphics[width = 0.32\textwidth]{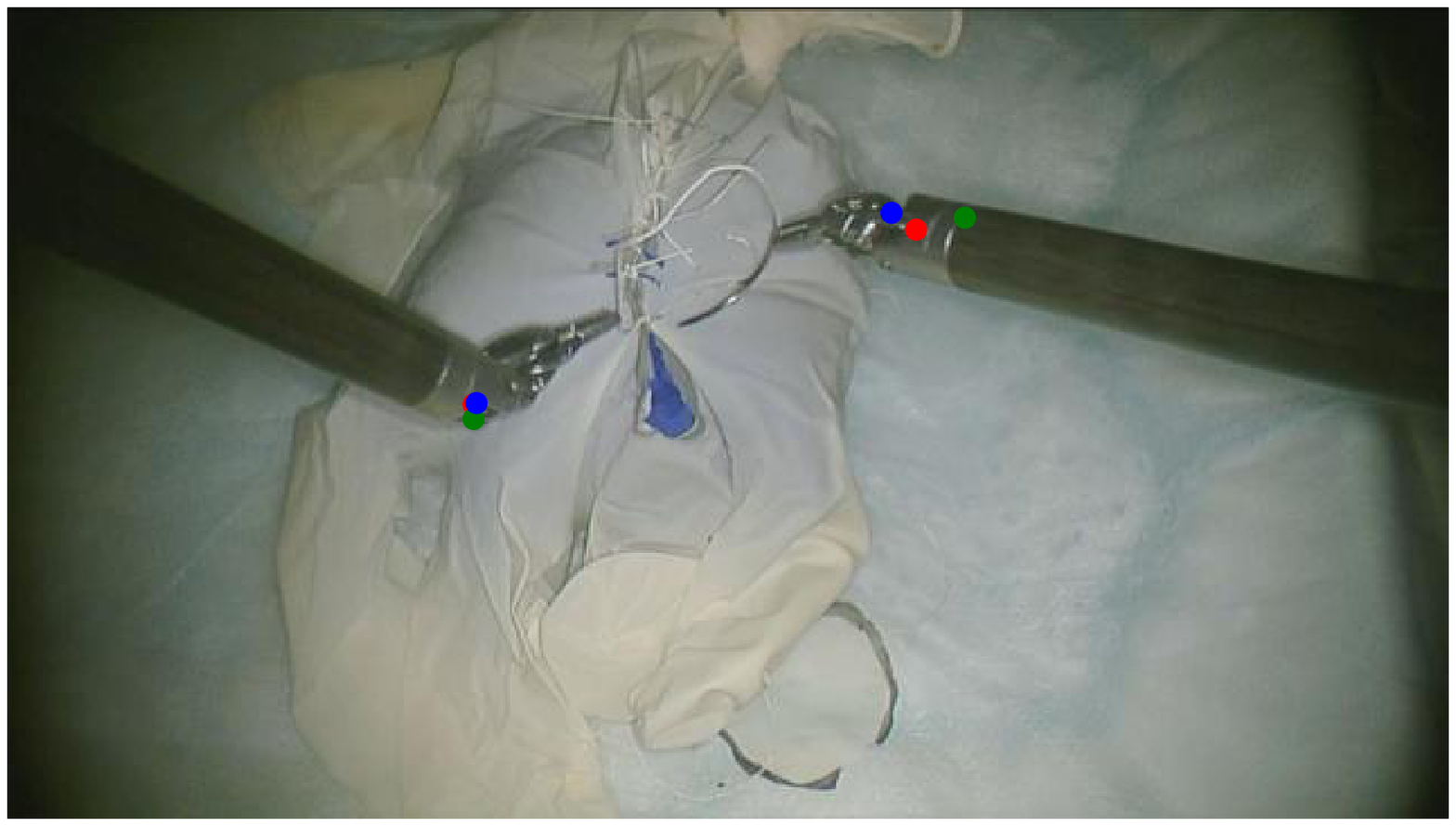}}
\subfigure[\label{fig:reproj-glove-2}]{\includegraphics[width = 0.32\textwidth]{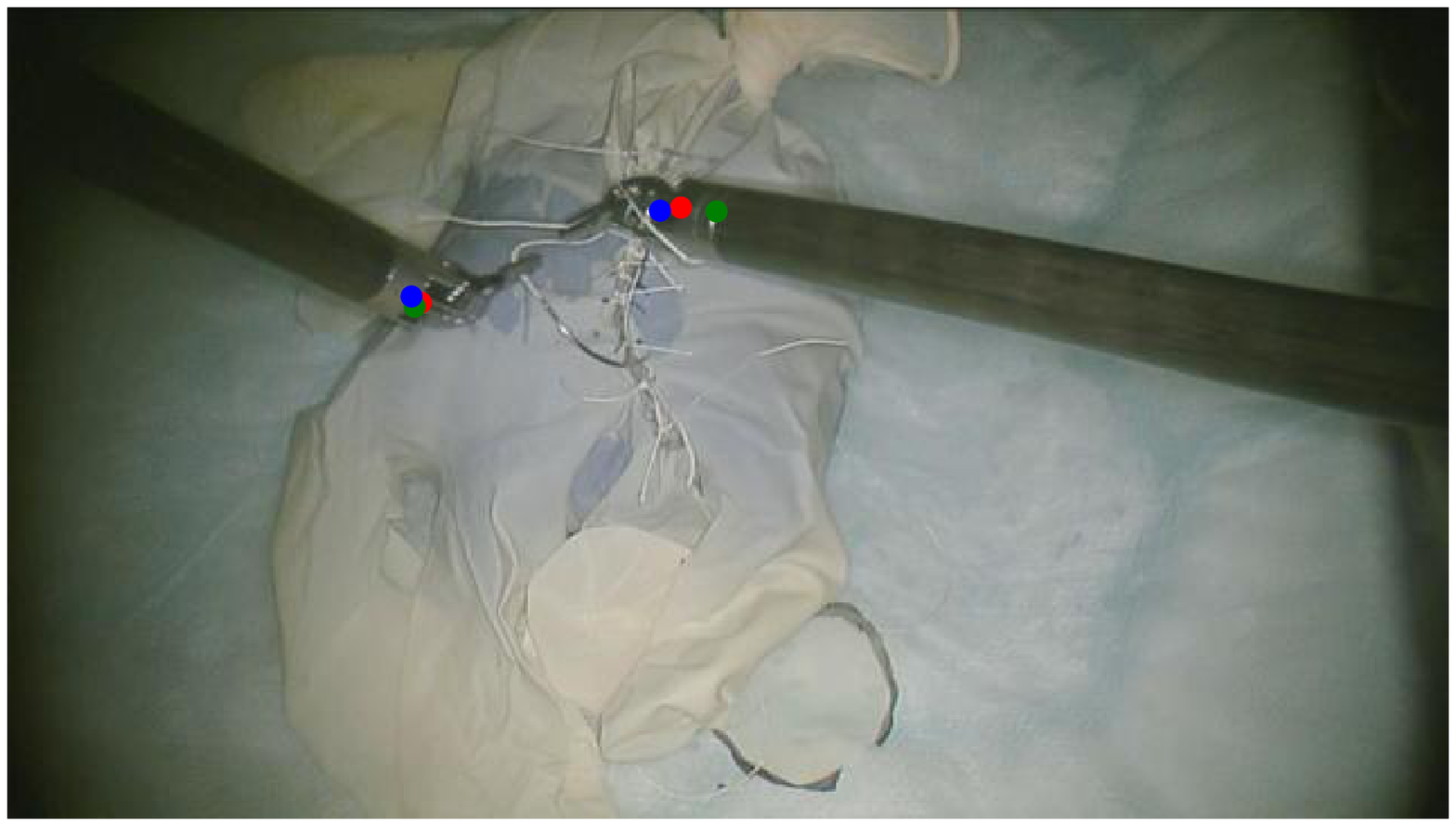}}
\subfigure[\label{fig:reproj-glove-3}]{\includegraphics[width = 0.32\textwidth]{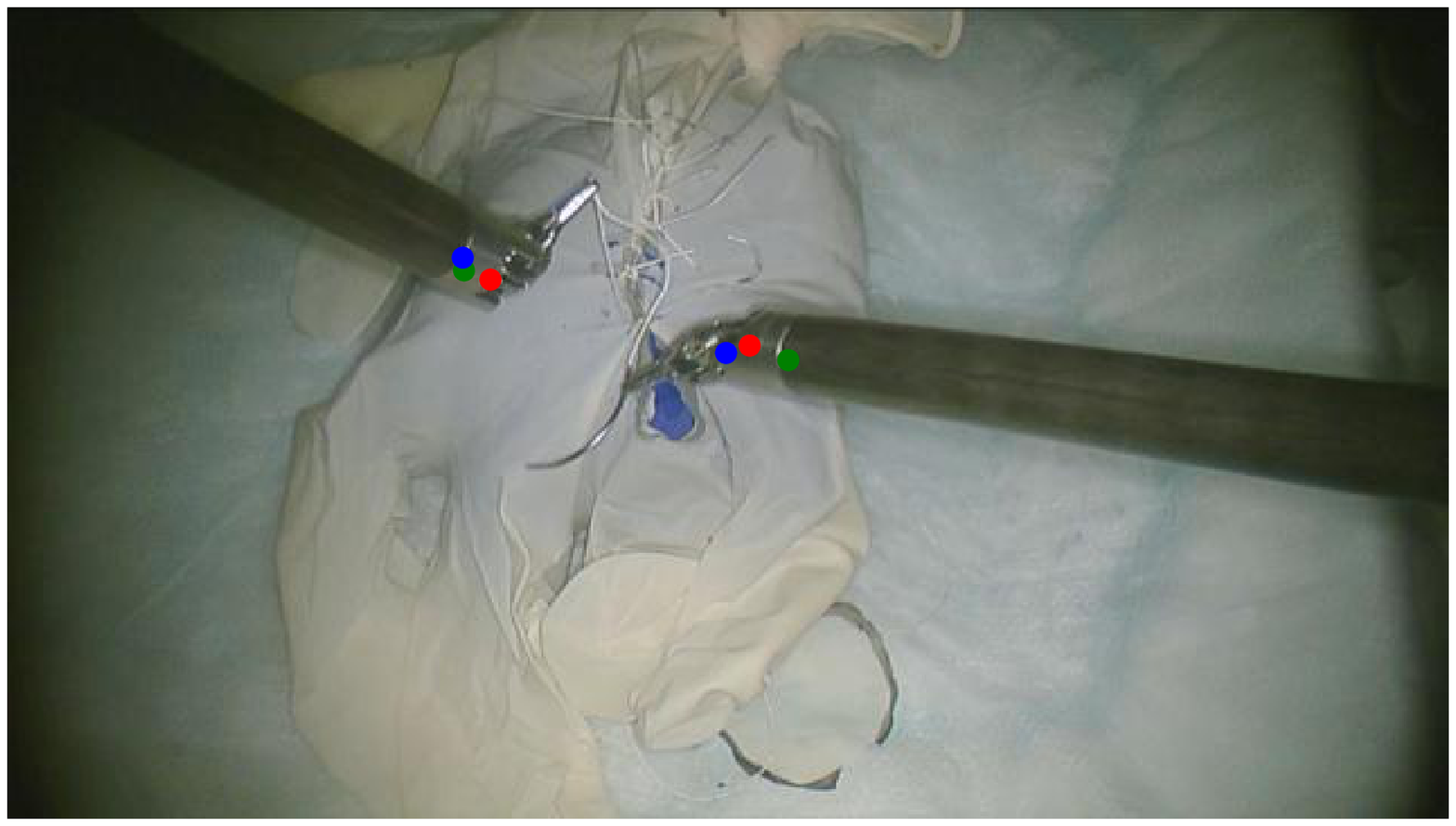}} \\
\subfigure[\label{fig:reproj-surg-1}]{\includegraphics[width = 0.32\textwidth]{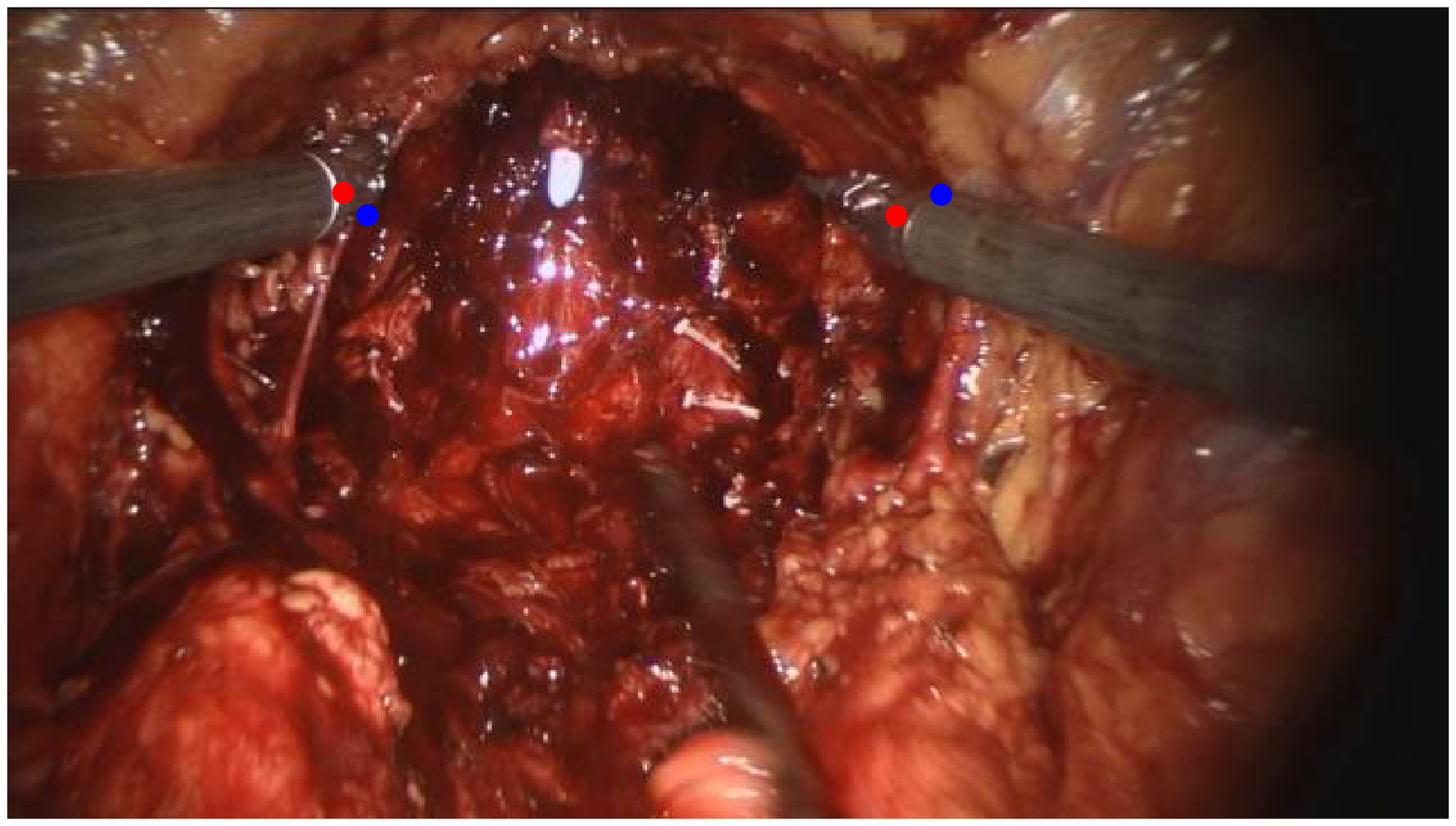}}
\subfigure[\label{fig:reproj-surg-2}]{\includegraphics[width = 0.32\textwidth]{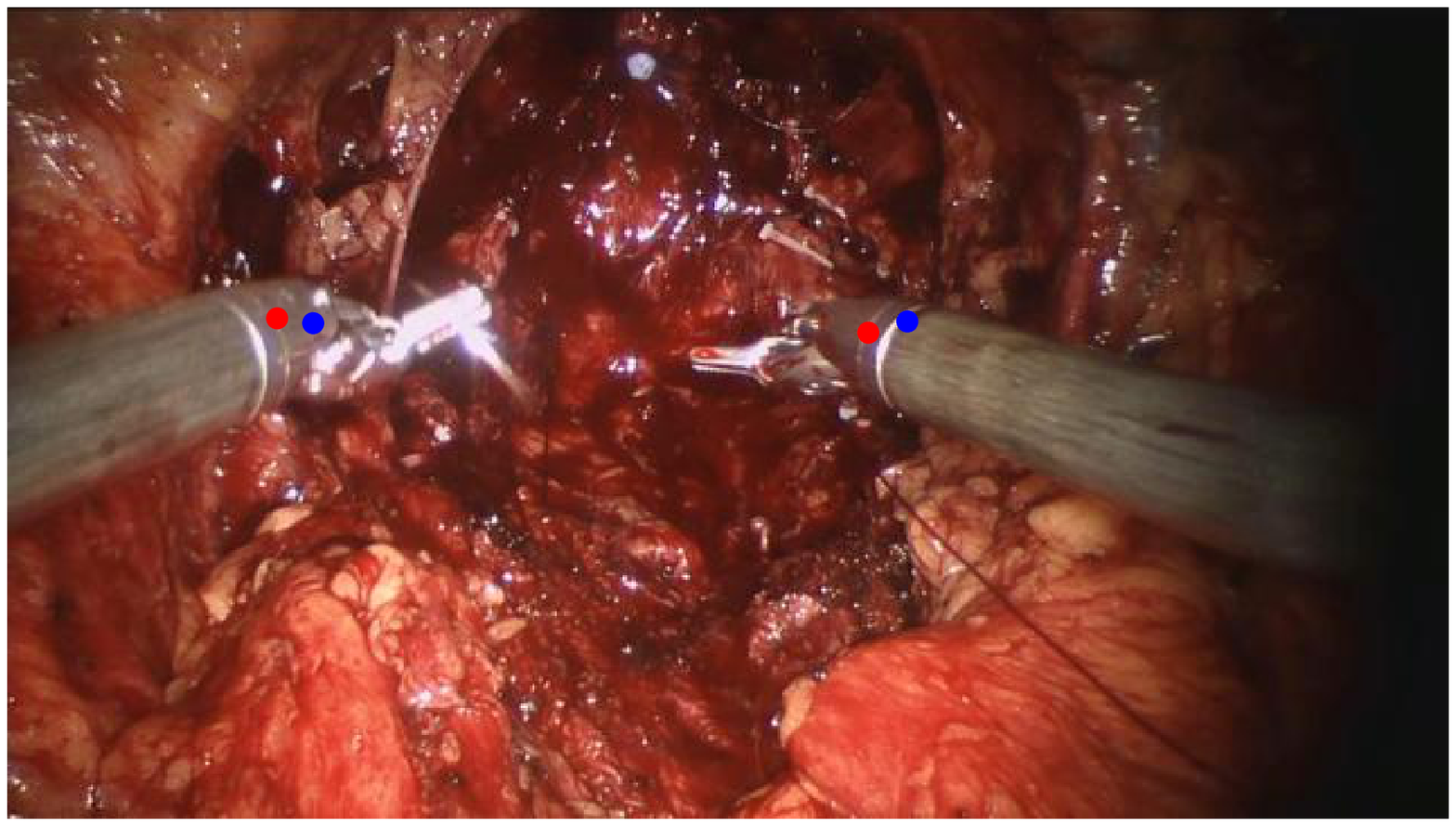}}
\subfigure[\label{fig:reproj-surg-3}]{\includegraphics[width = 0.32\textwidth]{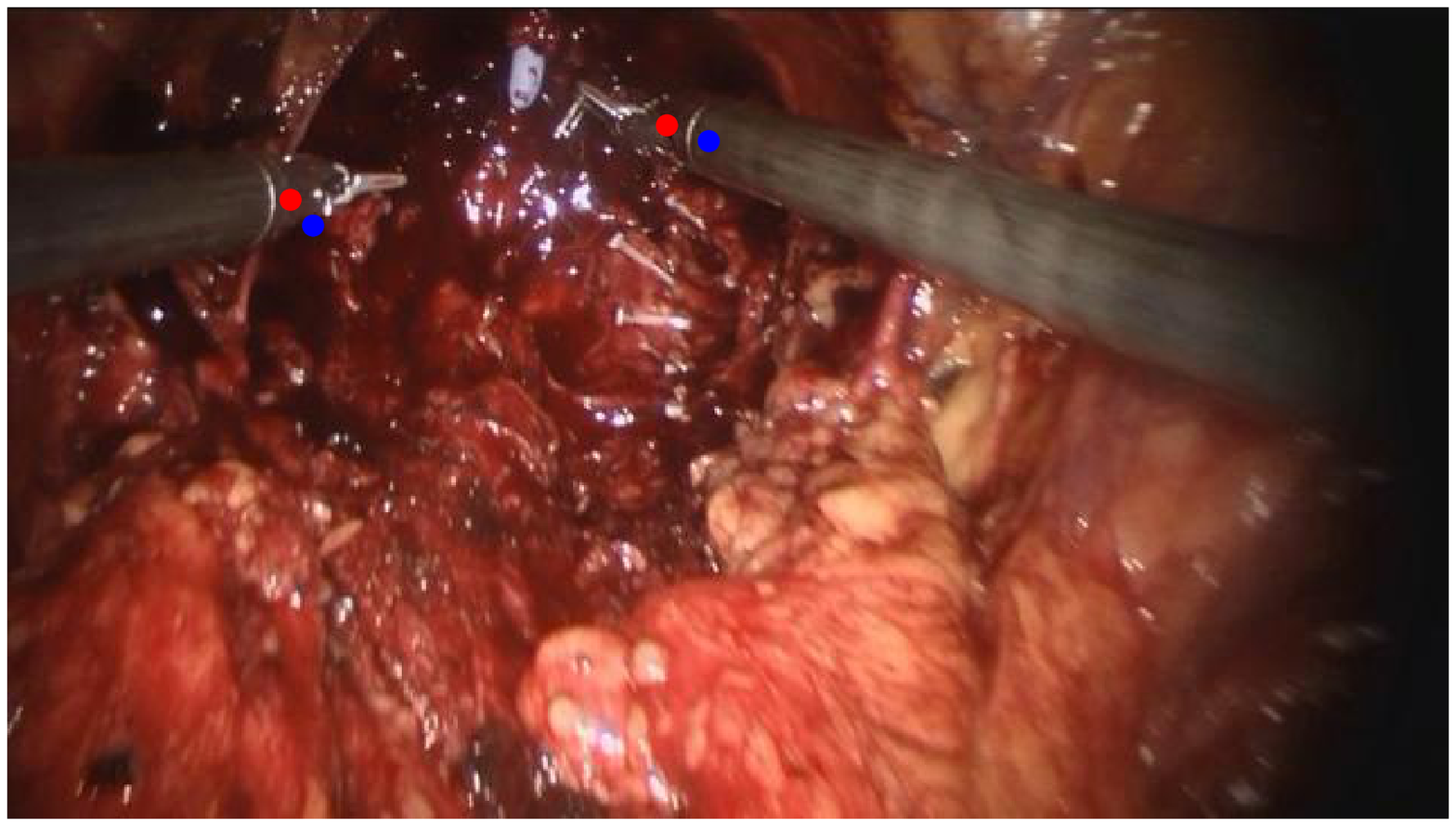}} \\ 
\subfigure[\label{fig:reproj-sens-1}]{\includegraphics[width = 0.32\textwidth]{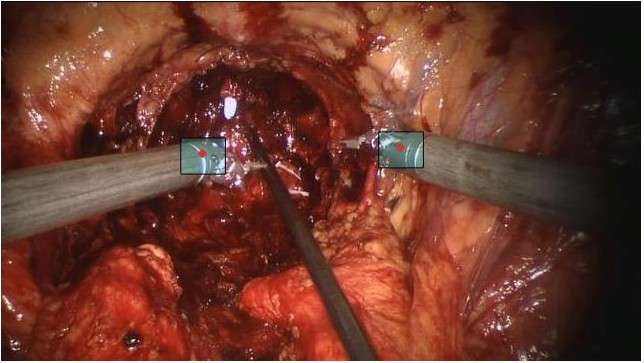}}
\subfigure[\label{fig:reproj-sens-2}]{\includegraphics[width = 0.32\textwidth]{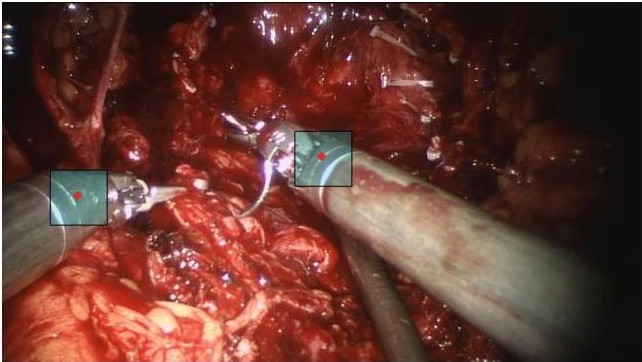}}
\subfigure[\label{fig:reproj-sens-3}]{\includegraphics[width = 0.32\textwidth]{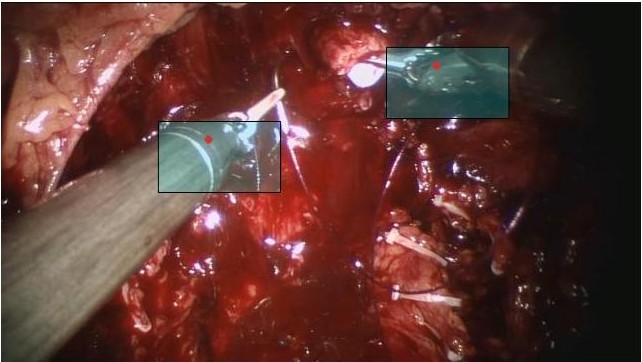}}  
\caption{The re-projected PSM1's PSM2's positions onto unseen images. PSM1 is the tool on the right and PSM2 is on the left. The red dots are the labelled position obtained from DLC package. The green and blue dots are the re-projected positions from using the method in \cite{9495280} and the newly proposed network, respectively. The differences between the two sets of data are the unaccounted features by the kinematic data in the glove suturing dataset ((a)-(c)) are much less than those in prostatectomy dataset ((d) - (f)) where it has specularity, blood, and occlusion, and the camera is constantly moved during prostatectomy. (g)-(i) The labelled positions are marked in red and cyan rectangulars show the range of re-projection error caused by 1 to 3 millimetres miscalibration. The range of re-projection error also depends on how close the camera is to the instruments. In this particular dataset, the error range is from 10 pixels to 40 pixels (as shown in (i)).}
\label{fig:result-label}
\end{figure*} 
\subsection{Calibration results}\label{subsec:exp}
To validate the calibration performance, we use the re-projection error as the metric to compare between two networks, as the ground truth of the hand-eye transformation is not known. The re-projected points in glove suturing and prostatectomy datasets are shown in Figure \ref{fig:reproj-glove-1}-\ref{fig:reproj-glove-3} and \ref{fig:reproj-surg-1}-\ref{fig:reproj-surg-3}, respectively. For the prostatectomy dataset, only the proposed network is validated, since the network in \cite{9495280} already fails the calibration and yields unusable results.

The results from glove suturing dataset show that the previous network yield accurate re-projected positions of the instrument with a few outliers. This is evident in the deviation in PSM1's re-projected positions. On the other hand, it is apparent that the new network gives a more accurate and consistent re-projection for both tools. Athough the camera is not moved in this dataset, the average re-projection error yielded by the proposed network are $16.88\pm6.12$ and $10.59\pm5.01$, whereas the previous network's is $19.67\pm7.39$ and $14.69\pm9.73$ pixels.

On the prostatectomy dataset, there are occlusions, specularity and non-linear errors in the kinematic chain caused by the movement of the camera which increases difficulty in calibrating the hand-eye parameters as shown in the previous paper \cite{9495280}, since most of them are not accounted by the kinematic. Therefore, the re-projected positions are not as close to the ones marked by DLC as they are in glove suturing dataset. The re-projection errors are $27.70\pm16.01$ and $21.83\pm12.86$ for PSM1's and PSM2's, respectively. This signifies that introducing Eq. \ref{eq:reproj-diff} to the loss function induces the change in the hand-eye parameters, and in turn compensates non-linear errors in the kinematic chain through the resulting transformation and improves the overall calibration performance.

Figure \ref{fig:reproj-sens-1}-\ref{fig:reproj-sens-3} show how a 1-2 millimetre-miscalibration can affect the re-projection error. In the scenes, as the camera moves closer to the tools, the range of re-projection becomes wider. It can be implied that although the network yields a rather high re-projection error in comparison to conventional camera or hand-eye calibration techniques, the calibration error obtained from the network is still no more than a couple of millemetre. Moreover, extensive study in the literature also shows that the state-of-the-art conventional hand-eye calibration approaches yield around 1-3 millmetres calibration error in the translation component. This shows that calibrating the hand-eye matrix using neural network can also achieve the similar accuracy with the potential of the procedure being online and more adaptive.

The evidence of the change in the calibration result corresponding to the camera movement is shown in Figure \ref{fig:result-label}. This is especially clear from the frame 200 and around 300 where the camera moves and stops and the estimation reacts accordingly. Although the link between the camera and the end-effector of ECM itself may not change during any motion when it is established, it can be argued that non-linear errors can occur at any points in the chain and result in a deviation in ECM's end-effector pose which require the compensation at the hand-eye parameters.
\begin{figure*}[!t]
\centering
\subfigure[\label{fig:handeye-converge-plot}]{\includegraphics[width = 0.32\textwidth]{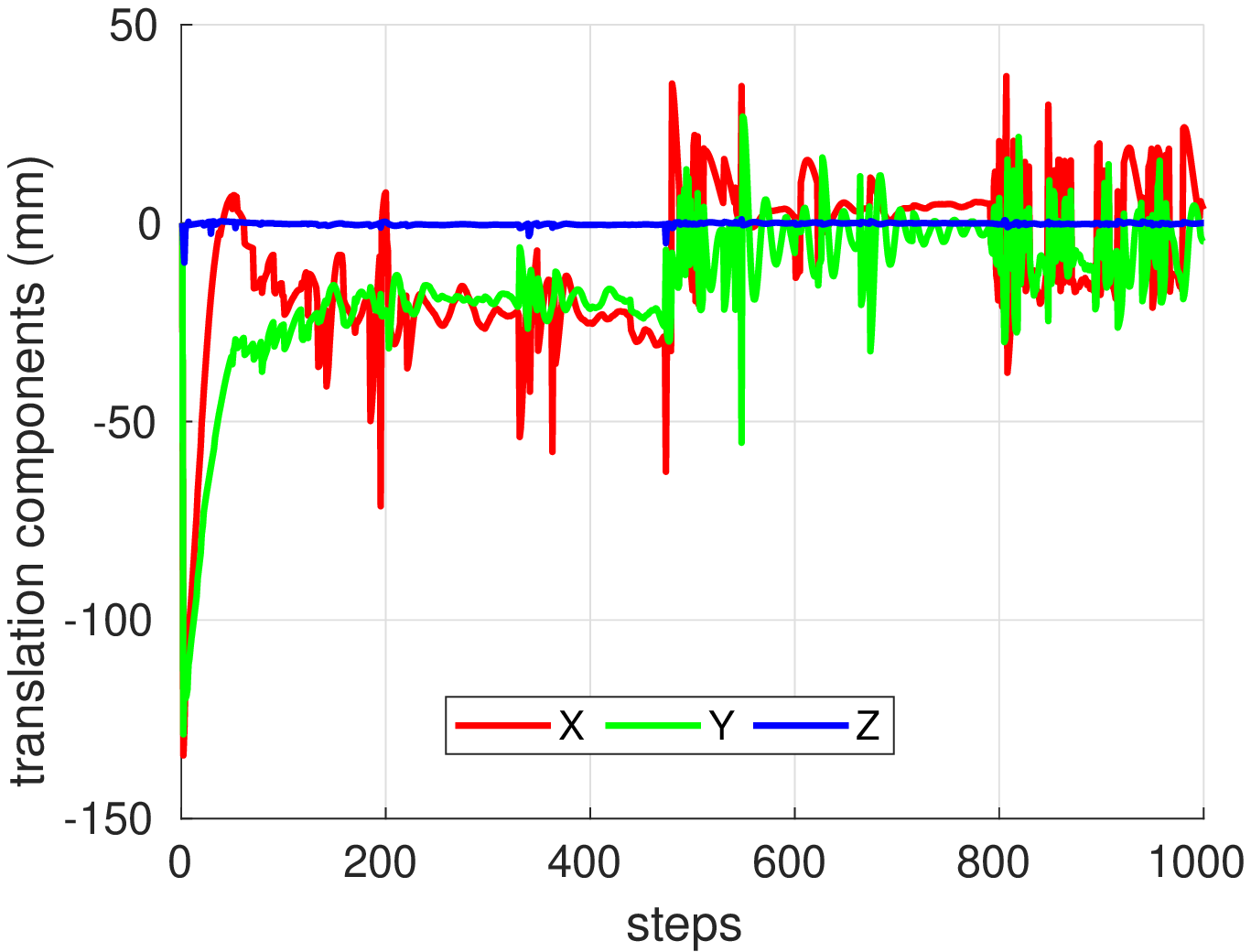}} 
\subfigure[\label{fig:train-converge-plot}]{\includegraphics[width = 0.32\textwidth]{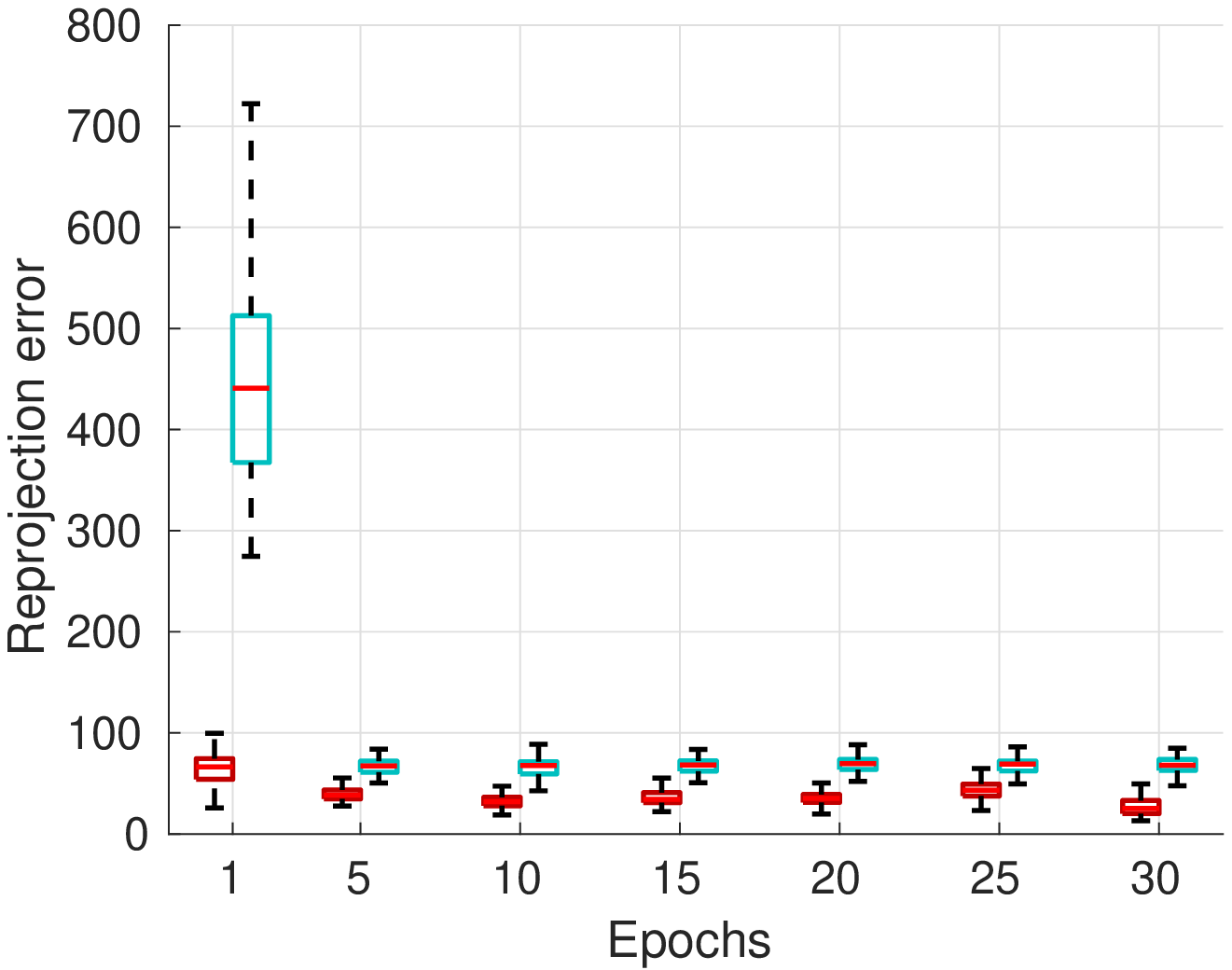}}
\caption{(a) How the translation component in the estimated hand-eye converges over the steps during the training within the first epoch. (b) The re-projection errors obtained from using the calibration results from different network (red: the proposed network, cyan: the previous network in \cite{9495280}).}
\label{fig:result-label}
\end{figure*} 

Furthermore, Figure \ref{fig:handeye-converge-plot} and \ref{fig:train-converge-plot} show that training using the proposed network gives a more accurate re-projection than the previous network, whereas in the previous version, it takes longer time to improve and yield less accuracy. This indicates that the proposed objective function (Eq. \ref{eq:reproj-diff} is an effective constraint to the neural network-based hand-eye problem in RMIS and the change caused by the equation also improves the estimation, as it is not only moving toward one pre-determined rigid transformation, but is rather induced to slightly change in every input and able to escape the local minima.
\subsection{Discussion}\label{subsec:disc}
The proposed method tackles two challenges in the hand-eye problems in RMIS. Firstly, the objective function does not rely on the original hand-eye equation $\mathbf{AX} = \mathbf{XB}$ which is proved to be ill-conditioned in the RMIS setup \cite{Huang2018OnGM}. It instead solves the problem by minimising the re-projection error. Secondly, the use of neural network significantly simplifies the calibration procedure as it no longer needs the setup of the calibration environment.

Regarding the noise in the calibration, it is shown in the literature that a small discrepancy in the kinematic chain or miscalibration of the camera parameters can result in a noticeable calibration error \cite{Pachtrachai2018}. The reason is that the hand-eye equations assume that the hand-eye transformation $\mathbf{X}$ is fixed and the solutions rely mainly on the kinematic chain and the camera movement where they can be erroneous when solving the equations. Furthermore, the error can be caused by unclear features and artefact which essentially can be propagated to the calibration result. 


Explicitly modelling this error in the chain is a challenging task as the error can be influenced by many external factors such as poses, weight of the attached tools, or mechanical compliance \cite{ZHANG2021102595}. Hence, in addition to advantages of the proposed neural network-based method, the introduced objective function induces the change in the hand-eye parameters over time, instead of fitting the estimated transformation to the pre-determined hand-eye. The purpose of this function is to compensate the errors occurring in the kinematic chain caused by the tendon driven actuators due to forces at the end-effector and joint compliance. The results show that the compensation indeed improves the calibration performance as can be seen in the more accurate re-projected instruments' positions.

Despite a significant improvement, as the calibration is a neural network-based approach, it still requires a tremendous amount of diverse dataset for the network to generalise. Currently, the network is trained with the glove suturing and the prostatectomy dataset and the network already yields a sensible re-projected positions, but it is likely to perform worse when presented with a different dataset containing unseen background or  operation. Therefore, to improve the calibration performance further, the network should be trained with a variation of dataset so that it can model the non-linear errors in the kinematic chain and uncertainty more accurately.
\section{Conclusion}\label{sec:conc}
This paper proposes a state-of-the-art deep network architecture to simplify and solve the hand-eye problem in RMIS. The method determines the hand-eye transformation from images, and kinematic data, unlike conventional methods where the calibration environment must be setup. The network requires LSTM to extract temporal features between data and DLC to mark the tools' positions for loss function. In addition to the previous formulation, we also introduce the derivative of the re-projection function to the loss to induce the change in the hand-eye parameters which compensate the non-linear errors in the kinematic chain. The results report a more accurate calibration performance in the glove suturing dataset where the camera is not moved and a sensible re-projected positions in the prostatectomy dataset, while the previous network fails to calibrate on this dataset. This indicates a significant improvement on neural network-based hand-eye methods and shows that unlike any general hand-eye setups in industry, accurate hand-eye calibration in RMIS requires a small deviation in the hand-eye parameters to compensate the overall kinematic errors. Future directions of this research includes the use of more features on the end-effector to give kinematic constraints on the re-projected points and bundle adjustment to further the re-projection error.

\section*{acknowledgement}
We would like to thank Intuitive Surgical, CA for providing us the da Vinci API and the support. The work was supported by the Wellcome/EPSRC Centre for Interventional and Surgical Sciences (WEISS) [203145Z/16/Z]; Engineering and Physical Sciences Research Council (EPSRC) [EP/P027938/1, EP/R004080/1, EP/P012841/1]; The Royal Academy of Engineering Chair in Emerging Technologies scheme.

\bibliographystyle{IEEEtran}
\bibliography{myBib}

\end{document}